# Detecting Volcano Deformation in InSAR using Deep learning

N. Anantrasirichai[1*], F. Albino[2], P. Hill[1], D. Bull[1], J. Biggs[2]

[1] *Visual Information Laboratory, University of Bristol, UK,*
[2] *Volcanology, University of Bristol, UK*
[*]*n.anantrasirichai@bristol.ac.uk*

Globally 800 million people live within 100 km of a volcano and currently 1500 volcanoes are considered active, but half of these have no ground-based monitoring. Alternatively, satellite radar (InSAR) can be employed to observe volcanic ground deformation, which has shown a significant statistical link to eruptions (Biggs, et al., 2014). Modern satellites provide large coverage with high resolution signals, leading to huge amounts of data. For example, the Sentinel-1 satellite allows us to acquire images of each of the world's volcanoes on a routine basis. It has a repeat cycle of 12 days and acquires data with a 250-km swath at a 5 m by 20 m spatial resolution (single look). This data is generated with greater than 10 TB per day or about 2 PB collected between 2014 and June 2017 (Fernández, et al., 2017). The explosion in data has brought major challenges associated with timely dissemination of information and distinguishing volcano deformation patterns from noise, which currently relies on manual inspection. Moreover, volcano observatories still lack expertise to exploit satellite datasets, particularly in developing countries.

Here, we present a novel approach to detect volcanic ground deformation automatically from InSAR images. This approach brings together satellite-based volcano geodesy and machine learning algorithms to develop new ways of automatically searching through large volumes of radar data to detect unusual patterns within the images. In this study, we use wrapped-phase InSAR images from NERC-COMET-LiCs (González, et al., 2016) covering volcanic regions in Ethiopia, Kenya, Italy and Turkey. Each image has an approximate size of 19500x19700 pixels (~1GB). A diagram of the proposed framework using deep learning with a convolutional neural network (CNN) is shown in Figure 1.

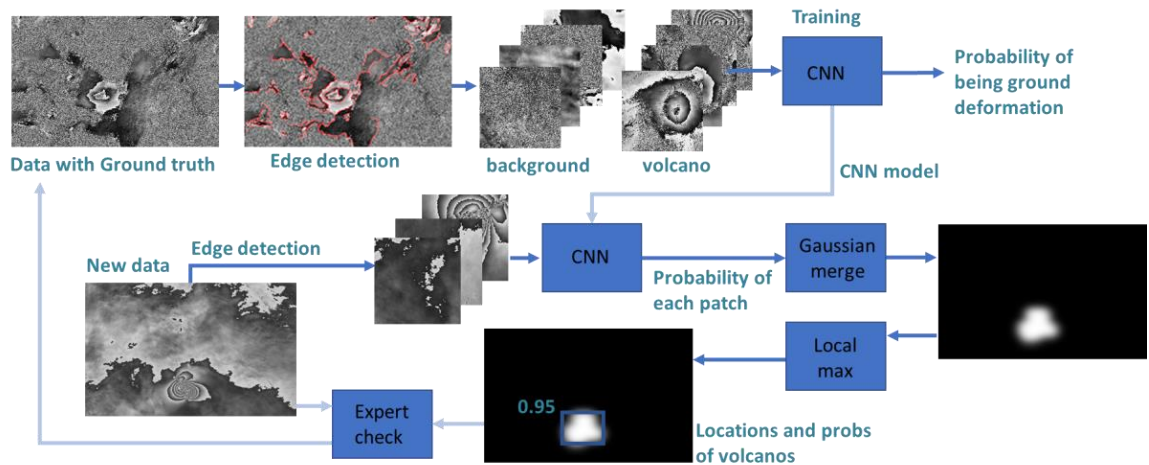

**Figure 1: A proposed method to detect volcano deformation**

Each training image is divided into patches equal to the size of input of the CNN (e.g. 224x224 pixels for AlexNet). They are overlapped by half of their size and the number of volcano patches are increased by shifting around the volcano area. As the number of background areas (negative samples) are significantly larger than those of the volcano patches (positive samples), only the background patches in which strong edges have been detected are used. To classify a potential signal as volcano deformation, patches with strong edges are tested and the probability results are merged with Gaussian weights. Finally, ground-truthing is performed by an expert and if the detection is a false positive, it is incorporated in the ground truth to update the CNN model.

Figure 2 shows the Receiver Operating Curves (ROC) for 2-fold cross validation for the considered dataset. It compares several pretrained CNN architectures and texture features with a support vector machine (SVM) classifier as a baseline. The best true-positive rate and true-negative rate are 0.899 and 0.992, respectively, which are the results of AlexNet. Generally big data presents significant challenges to deep learning because of its large scale, heterogeneity, and non-stationary distribution (Chen & Lin, 2014). Our results demonstrate that deep learning with CNNs (Krizhevsky, et al., 2012) has significant potential to capture characteristics of volcano deformation present in InSAR data. The next step is to test whether these methods are capable of distinguishing between deformation signals and atmospheric artefacts in single images, or whether full time series are required.

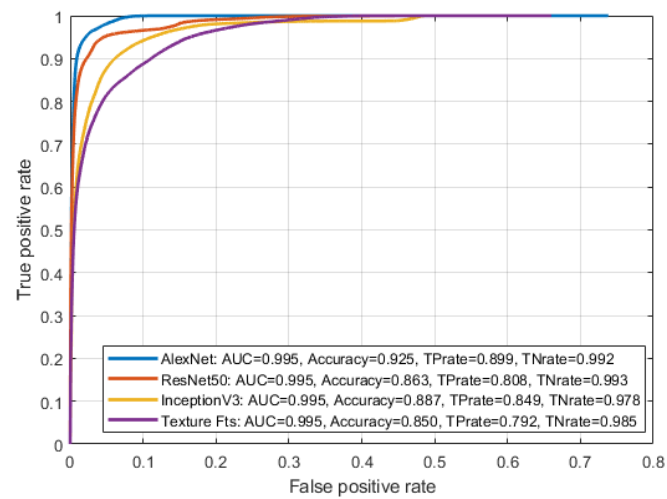

**Figure 2: ROC curves for the 2 folds of cross validation**

**References**
Biggs, J. et al., 2014. Global link between deformation and volcanic eruption quantified by satellite imagery. *Nature communications 5.*
Chen, X. W. & Lin, X., 2014. Big Data Deep Learning: Challenges and Perspectives. *IEEE Access,* Volume 2, pp. 514-525.
Fernández, J., Pepe, A., Poland, M. & Sigmundsson, F., 2017. Volcano Geodesy: Recent developments and future challenges. *Journal of Volcanology and Geothermal Research.*
González, P. et al., 2016. *LiCSAR: Tools for automated generation of Sentinel-1 frame interferograms.* s.l., AGU Fall Meeting.
Krizhevsky, A., Sutskever, I. & Hinton, G. E., 2012. ImageNet Classification with Deep Convolutional Neural Networks. *Advances in Neural Information Processing Systems,* Volume 25, pp. 1097-1105.